\documentclass[a4paper,fleqn]{cas-dc}

\usepackage{epstopdf}
\usepackage{xcolor,soul,framed}
\usepackage{array}
\usepackage{mdwmath}
\usepackage{mdwtab}
\usepackage{eqparbox}
\usepackage{url}
\usepackage{subcaption}
\usepackage{multirow}
\AppendGraphicsExtensions{.pdf}
\usepackage{caption}
\usepackage{subcaption}
\usepackage[numbers]{natbib}
\usepackage{soul}
\def\tsc#1{\csdef{#1}{\textsc{\lowercase{#1}}\xspace}}
\tsc{WGM}
\tsc{QE}
\tsc{EP}
\tsc{PMS}
\tsc{BEC}
\tsc{DE}

\begin{document}
\let\WriteBookmarks\relax
\def\floatpagepagefraction{1}
\def\textpagefraction{.001}
\shorttitle{Classification of ADHD Patients}
\shortauthors{Sartaj Ahmed Salman et~al.}

\title [mode = title]{Classification of ADHD Patients Using Kernel Hierarchical Extreme Learning Machine}                      
\tnotemark[1,2]

\tnotetext[2]{This paper supported in part ...}

\author[1]{Sartaj Ahmed Salman}[type=editor,
                        auid=000,bioid=1,
                        orcid=0000-0001-9344-6658]
\cormark[2]
\fnmark[1]
\ead{s2140019@edu.cc.uec.ac.jp}

\credit{Conceptualization of this study, Methodology, Software, Data curation, Writing - Original draft preparation}
\address[1]{Department of Informatics, The University of Electro-Communications (UEC), Chofu, Tokyo, Japan.}
\address[2]{School   of   Information   Management,  Nanjing  University (NJU),  Nanjing,  P.O.  Box  210023  P.R,  China.}
\address[3]{School of Computer Science and Engineering, Nanjing University of Science and Technology (NJUST), Nanjing, P.O. Box 210094 P.R, China.}

\author[3]{Zhichao Lian}
\cormark[1]
\fnmark[1,2]
\ead{lzcts@163.com}

\credit{MS.c. thesis supervision, project administration, funding acquisition}

\author%
[2]
{Milad Taleby Ahvanooey}
\credit{Conceptualization, formal analysis, review \& editing}
\ead{m.taleby@ieee.org}
\cortext[cor1]{Corresponding author}
\cortext[cor2]{Principal corresponding author}
\fntext[fn1]{S. A. Salman has completed this research during his MSc study at NJUST, PR. China in 2021}
\author[1]{Hiroki Takahashi}
\credit{Current PhD Advisor}
\ead{rocky@inf.uec.ac.jp}
\author[3]{Yuduo Zhang}
\ead{zyd@njust.edu.cn}

\credit{Lab-mate assists with software and data }

\begin{abstract}
Recently, the application of deep learning models to diagnose neuropsychiatric diseases from brain imaging data has received more and more attention. However, in practice, the exploration of interactions in brain functional connectivity based on operational magnetic resonance imaging data is critical for studying mental illness. Since Attention-Deficit and Hyperactivity Disorder (ADHD) is a type of chronic disease that is very difficult to diagnose in the early stages, it is necessary to improve the diagnosis accuracy of such illness using machine learning models treating patients before the critical condition. In this study, we utilize the dynamics of brain functional connectivity to model features from medical imaging data, which can extract the differences in brain function interactions between Normal Control (NC) and ADHD. To meet that requirement, we employ the Bayesian connectivity change-point model to detect brain dynamics using the local binary encoding approach and kernel hierarchical extreme learning machine for classifying features. To verify our model, we experimented with it on several real-world children datasets, and our results achieved superior classification rates compared to the state-of-the-art models.
\end{abstract}
\begin{keywords}
fMRI \sep BCCPM \sep Local features \sep Hierarchical Extreme Learning Machine \sep ADHD \sep Sparse-auto encoder
\end{keywords}
\maketitle
\section{Introduction}
These days, the Attention-Deficit and Hyperactivity Disorder (ADHD) is one of the most common neurodevelopmental disorders that influence up to 10\% of children all over the world, i.e., ADHD's symptoms often remain throughout the whole lifespan and affect to "comorbidities" same as the Major Depressive Disorder (MDD) \cite{RF1a}. Nevertheless, significant fluctuation of prevalence was appeared across different nations, considering diagnostic measures that are likely based on behavioral symptoms rather than impartial biomarkers. Moreover, because of the pathognomonic for multiple disorders, the ADHD's lack of biomarkers is particularly baleful since the overlap of main symptoms with other conventional psychiatric disorders such as personality, mood, and anxiety disorders. In addition, the diagnosis of ADHD for adults is obstructed by retrospective evaluation of symptoms during their childhood. Also, there exist disagreements between scientists/experts on ADHD as well as debates on misuse of ADHD treatment (e.g., Methylphenidate (MPH)) that have attracted research based on objective ADHD over subjective predictors with modest success so far \cite{RF2a}. \par
In the literature, there are three types of ADHD: combined, hyperactive/impulsive, and inattentive that have various symptoms when a patient is infected. The existing studies showed that ADHD is a disorder/disease, which happens in childhood and is classified based on symptoms such as difficulty in concentration, hyperactivity, short attention span, or impulsivity compared with children of the same age
\cite{RF1}. This diagnosis problem involves preliminary knowledge about an emergent disease and is challenging to identify because of insufficient information and physical symptoms.
Hence, diagnosing whether the
patient has ADHD or not was a hot topic for bioinformatics researchers.

Over the past decades, physicians or medical experts
have gathered information from different sources such as ADHD checklists \cite{RF3}. Moreover, the Neuro Bureau experts provide pre-processed data to contribute to the potential research for further improvements regarding the diagnosis of such disease (nitrc.org). Clinical studies have suggested that there are several illnesses, which alter the functional connectivity of the brain, and ADHD is also one of these disorders \cite{RF4}. In general, the human brain is the most complex system amongst all the other organs. While coordinating with the body, the brain regions continuously change their formation, and the areas that generate temporary correlation are assumed to be connected functionally \cite{RF5}. Hence, brain function connection modeling can better understand the pathological basis of neurological disorders. Building a network of brain function connections can describe the interactions between various functions of the brain \cite{RF6}. In the current literature, researchers have suggested several other neural network-based models for extracting disorder features from medical imaging data. But they mainly suffer from a low diagnosis accuracy rate due to the lack of insufficient considered features and variation of the functional connectivity changes in the brain.\par
Below, we summarize our contributions twofold.
\begin{itemize}
    \item We employ the bayesian connectivity change point Model to find the time of dynamic change of brain function interactions, enhancing the accuracy of extracted features from medical imaging data.
    \item We develop a deep learning model by applying the Local Binary Encoding Method for feature extraction and Kernel Hierarchical Extreme Learning Machine (KHELM) for classification.
\end{itemize}

\section{Related Works}
Recently, biomedical scientists have suggested several automatic diagnosis approaches for extracting mass features from Functional Magnetic Resonance Imaging (fMRI) data. For example, Waqas Majeed developed a new approach to assess that the reproducible spatiotemporal pattern of BOLD fluctuations is consistent with
previous research and may have vital information about brain activity at rest. It indicates that in the resting state, the brain is active. Moreover, several studies have considered the dynamics of brain
function connections \cite{RF8}. Lindquist\cite{RF9} have proposed a modified exponentially weighted moving average (EWMA) model, which can be applied to FMRI data, and then used it to analyze the change point of time series . Chang et al \cite{RF10} have investigated the dynamic connectivity of the brain signals by applying the
sliding-window methodology \cite{RF10}. In \cite{RF13}, Ren et al have recommended a dynamic
graph metrics strategy to characterize temporal changes of functional brain
networks. In \cite{RF7}, Atif Riaz et al have suggested a Hybrid
fMRI framework that utilizes affinity propagation clustering and density peak for functional connectivity. In \cite{RF34}, Ahmed et al. presented a classification model that and non-ADHD using ELM with different datasets. We have extracted different FC's using different time atlases and use the FC's for classification from our previous work we conclude that the accuracy of our model increases with the number of extracted FC.
During the past few years, researchers have introduced different types of models for the
classification of ADHD. As this illness is a medically brain disorders that physicians usually diagnose it by assessing some symptoms through a subjective observation process. Therefore, bioinformatial studies have identified this disorder as a two-category problem, such as ADHD and non-ADHD. Gülay Çıçek as well as created two separate datasets, including gray level co-occurrence matrix and Haralick texture features for classification purposes using the machine learning algorithms \cite{RF12}.

Recently, there have been many advances in this field, in \cite{RF35} Jie Wang uses fNIRS signals for functional connectivity and interval features for classification of ADHD and non-ADHD. Shuiqi Lui proposed a novel algorithm for ADHD classification based on (CDAE) convolutional denoising auto-encoder and (AdaDT) adaptive boosting decision trees\cite{RF36}. Yibin Tang et al. \cite{RF37} have introduced a self-encoding network with non-imaging fusion for ADHD classification, which achieves quite high accuracy. However, it has some limitations such as it does not work well with different datasets and is also not able to extract the required features from the fMRI data. 
Miao and Zhang \cite{RF13} have suggested a relief and VA -relief based feature extraction approach to achieve high precision classification. Later on, in 2017 Sudha et al. in \cite{RF14} have suggested a model to extract the gait signal characteristics of ADHD children from
the video signals, which provides the disease diagnosis and strengthen the cognition of sick children \cite{RF14}. Chang et al.\cite{RF15} have introduced a feature extraction method based on a texture point of view considering the isotropic local binary patterns on three orthogonal planes (LBP-TOP) that employ the support vector machine (SVM) to classify the identified features. Athena Taymourtash, use sparse based representation method by extracting the feature by cluster ICs and uses k-nn classifier to find out the EEG source differences between adults with ADHD and healthy controls \cite{RF16}. Later in \cite{RF17}, Zhang has introduced a the dual diagnosis model, which recognizes the feature space separation by applying sparse representation.. Juan L. Lopez Marcano explained that the United States allows using {the $\theta/\beta$} power ratio (TBPR) as a diagnostic feature of ADHD \cite{RF18}. F.M. Grisales-Franco uses a Dynamic Sparse Coding (DSC) method based on physiologically motivated Spatio-temporal constraints to construct non-stationary brain activity, they search the difference between ADHD and control groups using statistical results \cite{RF19}.

However, state-of-the-art models have studied ADHD diagnoses using different machine learning and deep learning models. Still, the literature suffers from an insufficient diagnosis rate which is a challenging task for scientists. Therefore, in this article, we propose a novel classification framework for enhancing the diagnosis accuracy the ADHD from the brain imaging data as a novel contribution to classifying children with and without ADHD. In particular, we apply the BCCPM to detect change points, and the local binary encoding method to extract local information from FIPs of the brain as well as employ the KH-ELM for classification.   

\begin{figure}[h!]
	\centering
		\includegraphics[scale=.06]{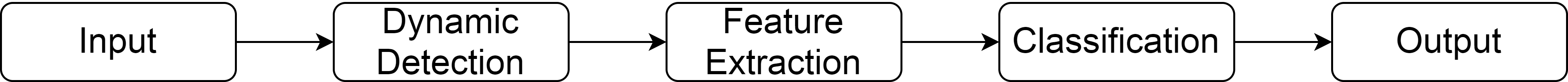}
	\caption{The data processing flow chart of the classification framework proposed in our paper}
	\label{FIG:1}
\end{figure}

\section{Proposed Models}
In this section, we describe two proposed models in details.
\subsection{Bayesian Connectivity Change Point Model}
Previous studies have shown that even when the brain is at resting status, it undergoes a series of dynamic changes, which affect the interaction of its functionalit connections. Iny this subsection, the proposed Bayesian connectivity change point model (BCCPM) for detecting dynamic changes in brain activities is explained. When the imaging data are classified directly without change point detection, the performance is not satisfactory. Therefore, we refer to the \cite{RF7} method to dynamically divide the time period into multiple time blocks, i.e., this process can actually identify the time points of dynamic brain change activity. Next, we will briefly introduce BCCPM.

With BCCPM, the data matrix $\ Z=(z_1,z_2,...,z_T)$ can be divided into several time blocks. The matrix $Z$ consists of $Tm$-dimensional vectors, where each vector represents the value of m ROIs (Regions of Interest) at time $t$. If each vector in $Z$ satisfies the independent and identical distribution, and $z_t \sim N(\mu,\sum)
t=1,2,...,T.$ 

Here, an indication vector$ L^\rightarrow =(L_1,L_2,...,L_T )$ is used, where $L_t=1$ represents the change point and the indication $L_1=1, L_t=0$ means unchanged. Then, the probability of $Z$ can be expressed as follows:
\begin{equation}
p(Z\mid L^\rightarrow) = \prod_{b=1}^{\sum L_i} p(Z_b)   
\end{equation}

where $\sum L_i$ is the number of divided time blocks, and $p(Z_b)$ can be obtained by the following equation (2). \cite{RF20}:
\begin{equation}
\begin{split}
p(z_1, z_2,...,z_T) =
\frac{p(z_1,z_2,...,z_T;\mu,\sum)}{p(\mu,\sum|z_1,z_2,...,z_T)}=\\
(\frac{1}{2\pi})^\frac{mT}{2}(\frac{k_0}{k_T})^\frac{m}{2}
\frac{\Gamma_m(\frac{v_T}{2}) (det
(\Lambda_0))^\frac{v_0}{2}}{\Gamma_m(\frac{v_0}{2}) (det
(\Lambda_T))^\frac{v_T}{2}} 2^\frac{mT}{2}
\end{split}
\end{equation}

where $ \Gamma_m $ is the multivariate gamma function.

Since $Z$ is divided into independent blocks of time, the posterior distribution of $p(L^\rightarrow|Z)$ is as follows:
\begin{equation}
    p(L^\rightarrow \mid Z)\propto p(L^\rightarrow)\cdot p(Z\mid L^\rightarrow )                       
\end{equation}

and $p(L^\rightarrow)=\sum_{t=1}^{T} p(L_t)$, where $p(L_t)\sim
Bern(0.5)$.Eventually, we employ the Markov chain Monte-Carlo \cite{RF21} with random $L^\rightarrow$ to generate samples with posterior distribution $p(L^\rightarrow\mid Z)$. See \cite{RF7} for details.
\subsection{Local Binary Encoding Method (LBEM)}
In order to extract the local information of brain function interaction patterns (FIPs), we proposed in our previous research that LBEM.
\cite{RF22} can be applied to local feature extraction from fMRI data.

The LBEM processes the data object DICCCOL-fMRI and the DICCCOL-fMRI is the data set matrix \[D=(D_1,D_2,...,D_T)\in R^{358*T}\], where each vector $D_t=(d_1,d_2,...,d_{358}),1 \leq t \leq T$ represents the value of 358 ROIs at time $t$.

\begin{figure}[h!]
	\centering
		\includegraphics[scale=.15]{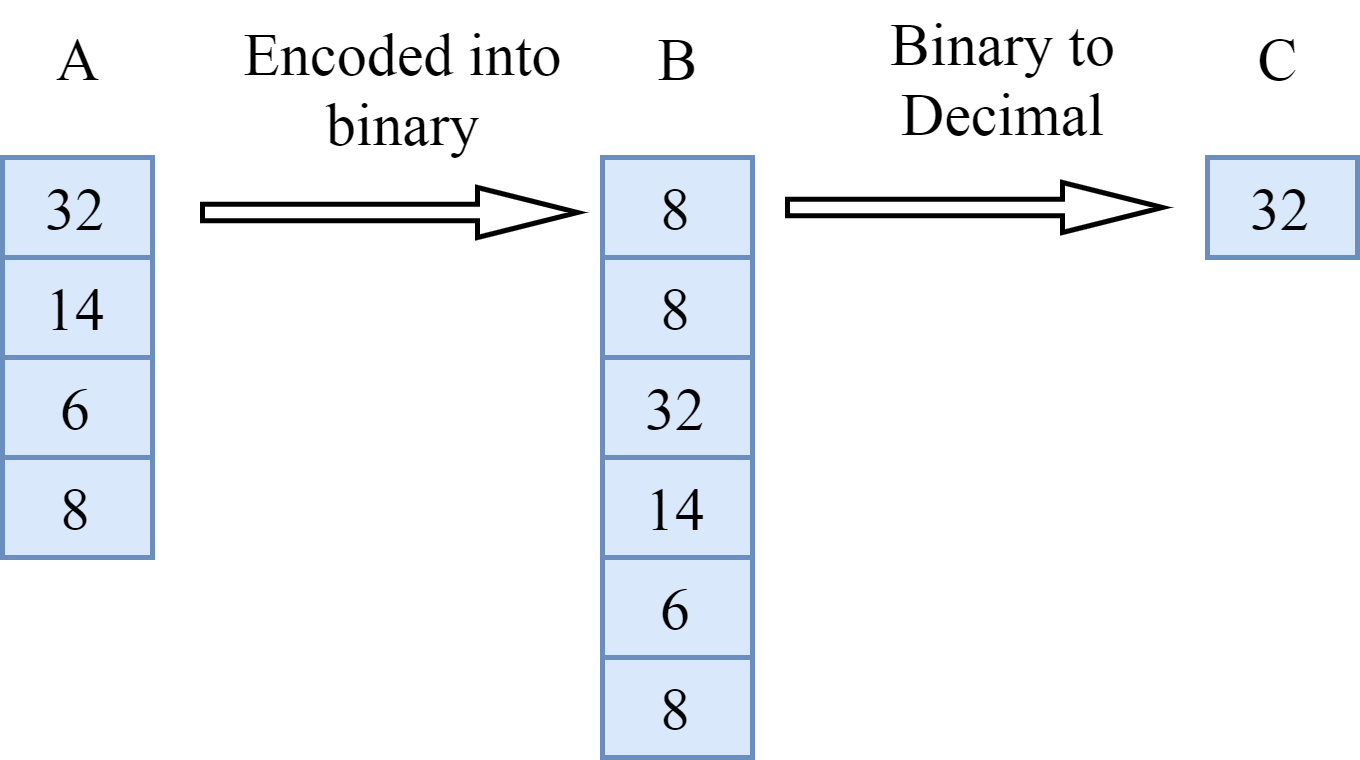}
	\caption{Encoding data from A to binary pattern B and then to decimal pattern C}
	\label{FIG:2}
\end{figure}

First, we consider a re-encoding process in each column vector $D_t$ within the data matrix $D$. If each element in $D_t$  is compared with its neighbouring elements, we have a new vector $F_t=(f_1,f_2,...,f_714)$, $1<t<T$ that can be denoted by the
equations (4-6):
\begin{equation}
    f_{2(i-1)-1}=\lbrace^{1,d_i\leq d_(i-1)}_{0,d_i>d_(i-1) },2\leq i \leq 358              
\end{equation}
\begin{equation}
    f_{2(i-1)}=\lbrace^{1,d_i\leq d_(i+1)}_{0,d_i>d_(i+1) },2\leq i \leq 357              
\end{equation}
\begin{equation}
    f_{724}=\lbrace^{1,d_{358}\leq d_1}_{0,d_{358}>d_1}              
\end{equation}
Then, we get a matrix of binary, which contains only 0 and 1 elements $F=(F_1,F_2,...,F_T)\in R^{714*T}$. Figure \ref{FIG:2} shows an example of a conversion to
binary matrix form. As shown, the vector$ A = (a_1,a_2,a_3,a_4)$ is encoded to obtain the vector $B =(b_1,b_2,...,b_6)$. Since $a_2<a_1$ assigns $b1$ to $1$, and $a_2>a_3$ assigns $b2$ to $0$. The values of $b_3,b_4,b_5$ can be determined in the same way, $b_6$ is equal to $1$, because of $a_4<a_1$ and the process of encoding into a binary would continue until
we obtain the remaining element of vector $F$. As a final assignment, all the Binary values of vector $B$
was converted in to decimal form.
Each of the six elements of the vector $F_t$ is assigned into a group. Next, each group number is converted from a 6-bit binary form to a decimal form, and the range of the converted decimal number is $[0, 63]$. After transforming the $F4$, we have the data matrix
$ E=(E_1,E_2,...,E_T)\in R^{119*T}$, which represents the local information between each ROI and its neighbouring ROIs. Figure \ref{FIG:2} depicts the example of encoding process.

\begin{figure*}
     \centering
     \begin{subfigure}[b]{0.3\textwidth}
         \centering
         \includegraphics[width=\textwidth]{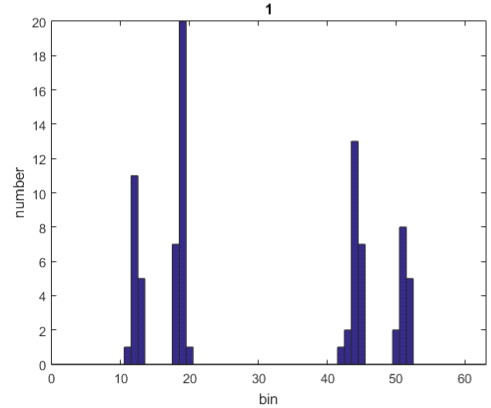}
         \caption{}
     \end{subfigure}
     \hfill
     \begin{subfigure}[b]{0.3\textwidth}
         \centering
         \includegraphics[width=\textwidth]{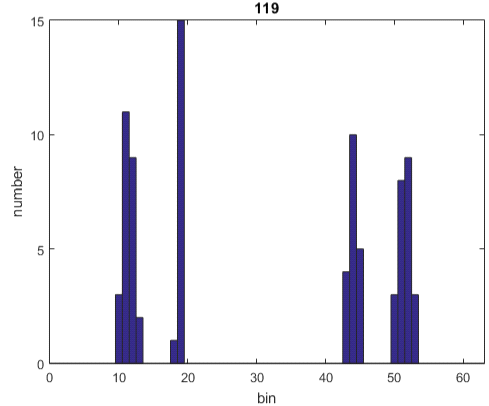}
         \caption{}
     \end{subfigure}
     \hfill
     \begin{subfigure}[b]{0.3\textwidth}
         \centering
         \includegraphics[width=\textwidth]{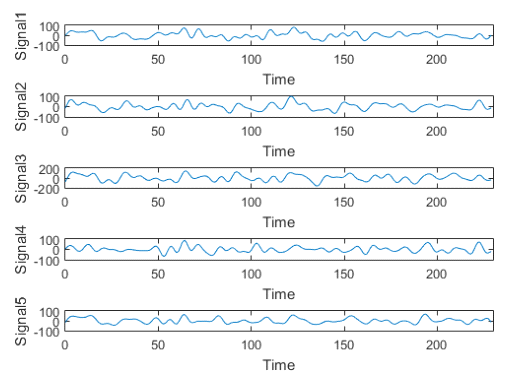}
         \caption{}
         \label{FIG:2a}
     \end{subfigure}
 \caption{(a,b) Number of discovered histograms (119), and (c) The r-fMRI signal of five DICCCOLs in one sample.}     
       
\end{figure*}

Third, we perform the histogram equalisation processing for each row vector of the data matrix $ E$ obtained in the second step. Thereby, the final output matrix is $O\in R^{119*64}$. As shown in Figure \ref{FIG:2a} (a,b), the sample object is processed to obtain 119 histograms.

\subsection{Kernel Extreme Learning Machine Algorithm (KELM)}

To extend the understanding of the KELM \cite{RF23}, we briefly explain the ELM \cite{RF24} theory, which only has one layer $H$ containing $L$ hidden nodes. Let us assume that the G is the classification problem, given a training set $\lbrace(x_j,z_j)\mid x_j\in R^d,z_j\in R^G\rbrace$, where $x_j$ is the vector, $z_j$ is the label, and $j=1,2,...,N$. The ELM model can be denoted:
\begin{enumerate}
  \item Calculate the hidden layer matrix H.
  \begin{equation}
        H= \left[ \begin{matrix}
h \left( x_{1} \right) \\
 \vdots \\
h \left( x_{N} \right) \\
\end{matrix}
 \right]  \left[ \begin{matrix}
h \left( w_{1}x_{1}+b_{1} \right)   &   \cdots   &  h \left( w_{L}x_{1}+b_{L} \right) \\
 \vdots   &  \ddots  &   \vdots \\
h \left( w_{1}x_{N}+b_{1} \right)   &   \cdots   &  h \left( w_{L}x_{N}+b_{L} \right) \\
\end{matrix}
 \right]
  \end{equation}
  
  where h ($ \cdot $ ) denotes the nonlinear activation function, the input weights w\textsubscript{i} and biases b\textsubscript{i} are randomly generated, $ i=1,2,  \ldots  , L.$
  \item The output weight vector $\beta$  can be calculate by equation (8).
  \begin{equation}
      \beta= H^\dag Z
  \end{equation}
  
  where $H^\dag$ is the Moore-Penrose generalized inverse of matrix $H$, and $Z$ is the training label matrix.
  \begin{equation}
      Z= \left[ \begin{matrix}
z_{1}^{T}\\
 \vdots \\
z_{N}^{T}\\
\end{matrix}
 \right] = \left[ \begin{matrix}
z_{11}  &   \cdots   &  z_{1G}\\
 \vdots   &  \ddots  &   \vdots \\
z_{N1}  &   \cdots   &  z_{NG}\\
\end{matrix}
 \right]
  \end{equation}
  
  The corresponding result of ELM can be obtained as follow:
  \begin{equation}
f \left( x_{j} \right) =h \left( x_{j} \right)  \beta        
  \end{equation}
  
   In KELM, a kernel matrix  \[ \Omega _{ELM}=HH^{T}: \Omega _{ELM}=h
   \left( x_{p} \right)  \cdot h \left( x_{q} \right) =K \left(
   x_{p},x_{q} \right)  \]  is applied. Then, the kernel case of ELM
   output function is:
   \begin{equation}
       ( f \left( x_{j} \right) = \left[ \begin{matrix}
K \left( x_{j},x_{1} \right) \\
 \vdots \\
K \left( x_{j},x_{N} \right) \\
\end{matrix}
 \right] ^{T} \left( \frac{1}{ \rho }+ \Omega _{ELM} \right) ^{-1}Z 
   \end{equation}
   
   In this work, a radial basis function as $\Omega_{ELM}$ is used to achieve better experimental results.
\end{enumerate}

\subsection{Kernel Hierarchical Extreme Learning Machine Algorithm (KH-ELM)}
\begin{figure*}
	\centering
		\includegraphics[scale=.3]{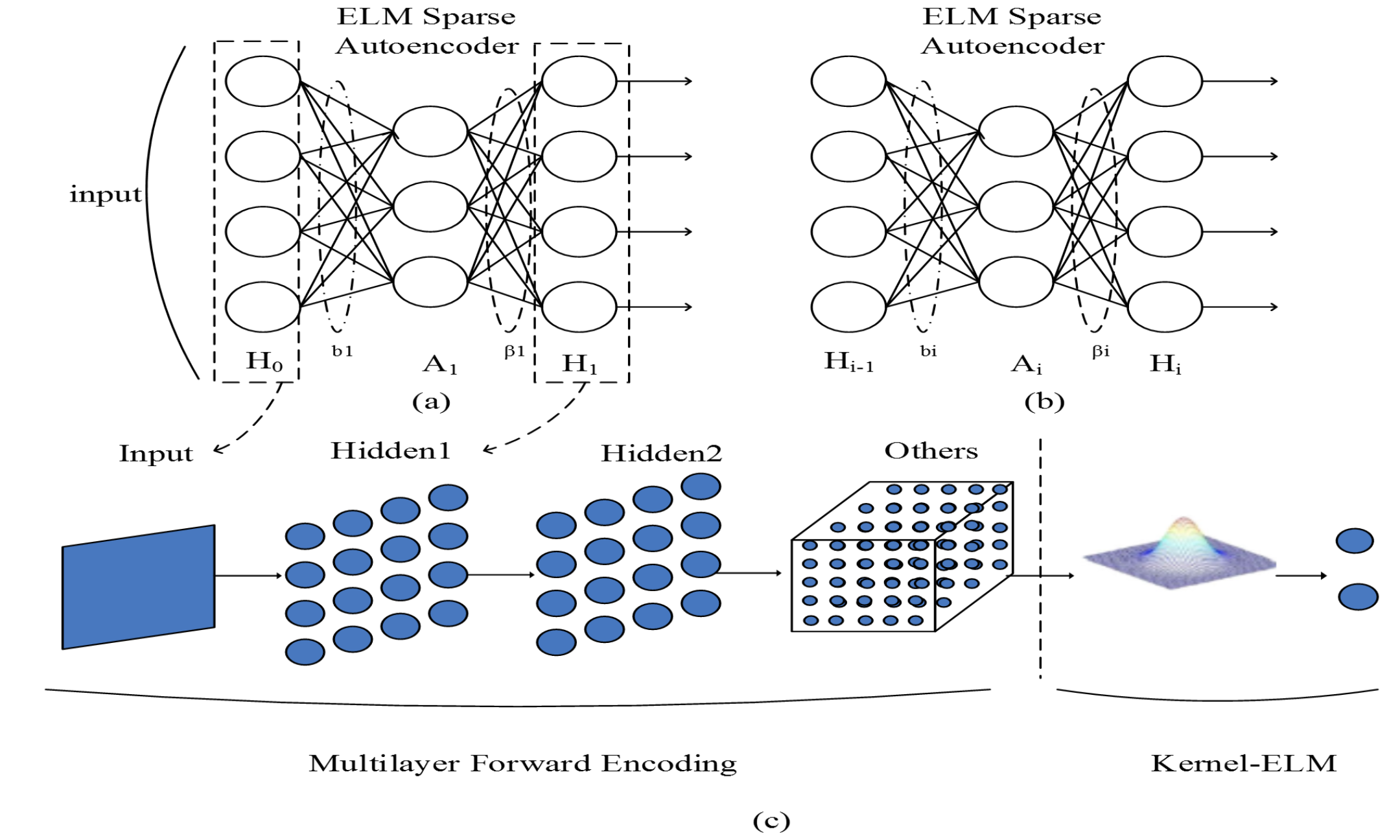}
	\caption{(a)The ELM Sparse Autoencoder structure of the first hidden layer. (b)The ELM Sparse Autoencoder structure of the ith hidden layer. (c)The architecture of the KH-ELM learning algorithm \cite{RF33}}
	\label{FIG:4}
\end{figure*}

The KH-ELM has a more complex structure than KELM, the training algorithm is different from the greedy lay-wise \cite{RF25}. Figure \ref{FIG:4} depicts the architecture of KH-ELM, which consists of two parts: \textit{i)} the unsupervised hierarchical feature extraction based on the ELM sparse-auto encoder, and \textit{ii)} the supervised classification based on KELM.

For the G-classification problem, given a training set.
\\\({(x_j,z_j)|x_j\in R^d,z_j\in R^G}\) where $x_j$ is the training vector, $z_j$ is the training label and $j=1,2,...,N$. Before unsupervised hierarchical feature extraction, the data-set should simply be scaled to $[0, 1]$ and the labels are adjusted to the G dimension, i.e. $-1$ or $1$. Then the input data is mapped into the ELM random feature space and extracted by n-layer unsupervised learning. For the KH-ELM learning algorithm, the input layer is taken as layer 0, the first layer is a hidden layer, the weights of the first layer $\beta$ are learned by the ELM sparse auto encoder (see Figure \ref{ FIG:4} (a)), and the other weights of the hidden layer are also learned by the ELM sparse-auto-encoder. The result of each hidden layer can be calculated as follows:
\begin{equation}
    H_{i}=g \left( H_{i-1} \cdot  \beta ^{T} \right) , 1 \leq i \leq n
\end{equation}

Where H\textsubscript{i} is the output of the ith layer, H\textsubscript{i-1} is the output of the (i-1)th layer, g( \( \cdot \) ) is the activation function of the hidden layers, and \( \beta \) is the weight of the ith hidden layer (output weight) \cite{RF26}. At KH-ELM, each hidden layer is an independent module. Once the feature of the previous hidden layer is extracted, the weights of the current hidden layer are set without fine-tuning \cite{RF27}. Here, the L1 regularisation norm is used to create a ELM auto-encoder so that the resulting ELM sparse auto-encoder provides better generalisation performance for the parameters. In addition, the use of penalty terms can constrain our model features. As a result, the learned model may have sparser features. The ELM sparse-auto encoder equation for this optimisation model is expressed as follows:
\begin{equation}
    ( O_{ \beta }=argmin_{ \beta } \{  \vert  \vert A \beta -X \vert  \vert ^{2}+ \vert  \vert  \beta  \vert  \vert _{l1} \}
\end{equation}

Here, $A$ denotes the random mapping output, \( \beta \)\, which is the weight of the hidden layer to be obtained, X represents the input data. In the proposed auto-encoder, A is a randomly initialised output mapped by a random weight matrix \( b= \left[ b_{1},b_{2}, \ldots , b_{n} \right] \) that does not require any optimisation. The structure of the ELM sparse-auto-encoder is shown in Figure \ref{ FIG:4} (a,b). As shown, this not only saves training time but also improves learning accuracy.

Next, we describe the ELM optimisation algorithm based on the sparse auto-encoder. In addition, we applied the Fast Iterative Shrinkage Threshold algorithm (FISTA) to solve the constraint minimisation problem for continuously differential functions. We use a FISTA type with constant steps. The specific algorithm is express as follows:
\begin{enumerate}
\item Calculate the Lipschitz constant $\gamma$ of the gradient $\nabla_p$ of the function $||A\beta-X||^2$ 
\item  Calculate $\beta$ iteratively. Begin the iteration by taking $y_1=\beta_0\in R^n$, $t1=1$ as the initial points. Then, for $i (i\geq 1)$

\ \ \ \ \ \ \ \ \ \ \ \  \textbf{Step 1.}  \(  \beta _{i}=S_{ \gamma } \left( y_{i} \right)  \) \textit{,} Where

\textit{  \( S_{ \gamma }=arg\min _{ \beta } \{ \frac{ \gamma }{2} \vert  \vert  \beta - \left(  \beta _{i-1}-\frac{1}{ \gamma } \left(  \beta _{i-1} \right)  \right)  \vert  \vert ^{2}+ \vert  \vert  \beta  \vert  \vert _{l1} \}  \) }

\ \ \ \ \ \ \ \ \ \ \ \ \  \textbf{Step 2.}  \( t_{i+1}=\frac{1+\sqrt[]{1+4t_{i}^{2}}}{2} \)

\ \ \ \ \ \ \ \ \ \ \ \ \  \textbf{Step 3.}  \( y_{i+1}= \beta _{i}+ \left( \frac{t_{i-1}}{t_{i+1}} \right)  \left(  \beta _{i}- \beta _{i-1} \right)  \) 

\end{enumerate}

The second part of KH-ELM functions based on the supervised feature classification employing the KELM in which the input is the matrix \( H_{n} \), and the output is the n-layer ELM sparse autoencoder. 

\section{Experimental Results and Analysis}
In this section, we conducted a comparative analysis by performing our model on the r-fMRI dataset, i.e., it \cite{RF6} contains 23 ADHD patients and 45 NC. To verify our proposed model, first, we selected 358 landmarks \cite{RF28} as nodes of the network. Second, we determined the DICCCOL for 68 subjects exploiting a commonly available tool in \cite{RF29}, and extracted the r-fMRI signal corresponding to it. Figure \ref{FIG:2a} shows the r-fMRI signal from five in one sample. After preprocessing, we obtain a $358*270$ time series matrix for each sample. Then, the r-fMRI time series of each sample was dynamically detected using BCCPM to obtain a $358*230$ time series matrix.

In this paper, 5-fold cross-validation is used to evaluate the proposed method. The experiment was repeated 30 times and then the average classification accuracy of each stage was calculated. To analogize via different approaches, the experimental results of three distinct methods were used: our proposed method along KH-ELM and KELM with the Jinli Ou \cite{RF6}, the KELM with different kernels and  of KH-ELM with multiple layers. First, it shows that our methodology with KELM achieves the better performance (the average of ADHD is 95.51\% and NC is 97.57\%), compared to the results of Ous \cite{RF6} (ADHD 92\% and NC 95\%). In addition, KELM has achieved more stable results than the Jinli Ou's technique. The comparison flash that it is effective to extract local features with LBEM. Then, it shows that the precision of KH-ELM, which is shown in the first column of table \ref{Tbl1} table. (97.17\% for ADHD and 99.76\% for NC) is higher than that of KELM. This expose that by using the ELM sparse auto-encoder to extract features from the training and the test dataset before KELM improves performance of the model.
\begin{table*}[h!]
\caption{Accuracy comparison of our proposed methods and Jinli Ou's method}
 			\centering
\begin{tabular}{p{0.33in}p{0.3in}p{0.3in}p{0.29in}p{0.29in}p{0.29in}p{0.29in}p{0.29in}p{0.29in}}
\hline
\multicolumn{1}{|p{0.33in}}{\multirow{1}{*}{\begin{tabular}{p{0.33in}}{\fontsize{8pt}{9.6pt}\selectfont \textbf{Class}}\\\end{tabular}}} & 
\multicolumn{2}{|p{0.33in}}{\multirow{1}{*}{\begin{tabular}{p{0.33in}}{\fontsize{8pt}{9.6pt}\selectfont \textbf{KHELM}}\\\end{tabular}}} & 
\multicolumn{2}{|p{0.33in}}{\multirow{1}{*}{\begin{tabular}{p{0.33in}}{\fontsize{8pt}{9.6pt}\selectfont \textbf{KELM}}\\\end{tabular}}} & 
\multicolumn{2}{|p{\dimexpr0.79in+2\tabcolsep\relax}} {\fontsize{8pt}{9.6pt}\selectfont \textbf{KELM without BCCPM}} & 
\multicolumn{2}{|p{\dimexpr0.79in+2\tabcolsep\relax}|}{{\fontsize{8pt}{9.6pt}\selectfont \textbf{Jinli Ou's Method}}} \\
\hline
\multicolumn{1}{|p{0.33in}}{} & 
\multicolumn{1}{|p{0.3in}}{{\fontsize{8pt}{9.6pt}\selectfont \textbf{\textit{NC}}}} & 
\multicolumn{1}{|p{0.3in}}{{\fontsize{8pt}{9.6pt}\selectfont \textbf{\textit{ADHD}}}} & 
\multicolumn{1}{|p{0.29in}}{{\fontsize{8pt}{9.6pt}\selectfont \textbf{\textit{NC}}}} & 
\multicolumn{1}{|p{0.29in}}{{\fontsize{8pt}{9.6pt}\selectfont \textbf{\textit{ADHD}}}} & 
\multicolumn{1}{|p{0.29in}}{{\fontsize{8pt}{9.6pt}\selectfont \textbf{\textit{NC}}}} & 
\multicolumn{1}{|p{0.29in}}{{\fontsize{8pt}{9.6pt}\selectfont \textbf{\textit{ADHD}}}} & 
\multicolumn{1}{|p{0.29in}}{{\fontsize{8pt}{9.6pt}\selectfont \textbf{\textit{NC}}}} & 
\multicolumn{1}{|p{0.29in}|}{{\fontsize{8pt}{9.6pt}\selectfont \textbf{\textit{ADHD}}}} \\
\hline
\multicolumn{1}{|p{0.33in}}{{\fontsize{8pt}{9.6pt}\selectfont Fold1}} & 
\multicolumn{1}{|p{0.3in}}{{\fontsize{8pt}{9.6pt}\selectfont 0.9973}} & 
\multicolumn{1}{|p{0.3in}}{{\fontsize{8pt}{9.6pt}\selectfont 1}} & 
\multicolumn{1}{|p{0.29in}}{{\fontsize{8pt}{9.6pt}\selectfont 0.9747}} & 
\multicolumn{1}{|p{0.29in}}{{\fontsize{8pt}{9.6pt}\selectfont 0.9522}} & 
\multicolumn{1}{|p{0.29in}}{{\fontsize{8pt}{9.6pt}\selectfont 0.7111}} & 
\multicolumn{1}{|p{0.29in}}{{\fontsize{8pt}{9.6pt}\selectfont 0.44}} & 
\multicolumn{1}{|p{0.29in}}{{\fontsize{8pt}{9.6pt}\selectfont 0.8889}} & 
\multicolumn{1}{|p{0.29in}|}{{\fontsize{8pt}{9.6pt}\selectfont 1}} \\
\hline
\multicolumn{1}{|p{0.33in}}{{\fontsize{8pt}{9.6pt}\selectfont Fold2}} & 
\multicolumn{1}{|p{0.3in}}{{\fontsize{8pt}{9.6pt}\selectfont 0.9907}} & 
\multicolumn{1}{|p{0.3in}}{{\fontsize{8pt}{9.6pt}\selectfont 1}} & 
\multicolumn{1}{|p{0.29in}}{{\fontsize{8pt}{9.6pt}\selectfont 0.9779}} & 
\multicolumn{1}{|p{0.29in}}{{\fontsize{8pt}{9.6pt}\selectfont 0.9449}} & 
\multicolumn{1}{|p{0.29in}}{{\fontsize{8pt}{9.6pt}\selectfont 0.7778}} & 
\multicolumn{1}{|p{0.29in}}{{\fontsize{8pt}{9.6pt}\selectfont 0.4}} & 
\multicolumn{1}{|p{0.29in}}{{\fontsize{8pt}{9.6pt}\selectfont 0.8889}} & 
\multicolumn{1}{|p{0.29in}|}{{\fontsize{8pt}{9.6pt}\selectfont 0.8}} \\
\hline
\multicolumn{1}{|p{0.33in}}{{\fontsize{8pt}{9.6pt}\selectfont Fold3}} & 
\multicolumn{1}{|p{0.3in}}{{\fontsize{8pt}{9.6pt}\selectfont 1}} & 
\multicolumn{1}{|p{0.3in}}{{\fontsize{8pt}{9.6pt}\selectfont 0.9974}} & 
\multicolumn{1}{|p{0.29in}}{{\fontsize{8pt}{9.6pt}\selectfont 0.983}} & 
\multicolumn{1}{|p{0.29in}}{{\fontsize{8pt}{9.6pt}\selectfont 0.959}} & 
\multicolumn{1}{|p{0.29in}}{{\fontsize{8pt}{9.6pt}\selectfont 0.6222}} & 
\multicolumn{1}{|p{0.29in}}{{\fontsize{8pt}{9.6pt}\selectfont 0.4}} & 
\multicolumn{1}{|p{0.29in}}{{\fontsize{8pt}{9.6pt}\selectfont 1}} & 
\multicolumn{1}{|p{0.29in}|}{{\fontsize{8pt}{9.6pt}\selectfont 0.8}} \\
\hline
\multicolumn{1}{|p{0.33in}}{{\fontsize{8pt}{9.6pt}\selectfont Fold4}} & 
\multicolumn{1}{|p{0.3in}}{{\fontsize{8pt}{9.6pt}\selectfont 1}} & 
\multicolumn{1}{|p{0.3in}}{{\fontsize{8pt}{9.6pt}\selectfont 0.9564}} & 
\multicolumn{1}{|p{0.29in}}{{\fontsize{8pt}{9.6pt}\selectfont 0.9711}} & 
\multicolumn{1}{|p{0.29in}}{{\fontsize{8pt}{9.6pt}\selectfont 0.9463}} & 
\multicolumn{1}{|p{0.29in}}{{\fontsize{8pt}{9.6pt}\selectfont 0.6667}} & 
\multicolumn{1}{|p{0.29in}}{{\fontsize{8pt}{9.6pt}\selectfont 0.55}} & 
\multicolumn{1}{|p{0.29in}}{{\fontsize{8pt}{9.6pt}\selectfont 1}} & 
\multicolumn{1}{|p{0.29in}|}{{\fontsize{8pt}{9.6pt}\selectfont 1}} \\
\hline
\multicolumn{1}{|p{0.33in}}{{\fontsize{8pt}{9.6pt}\selectfont Fold5}} & 
\multicolumn{1}{|p{0.3in}}{{\fontsize{8pt}{9.6pt}\selectfont 1}} & 
\multicolumn{1}{|p{0.3in}}{{\fontsize{8pt}{9.6pt}\selectfont 0.9048}} & 
\multicolumn{1}{|p{0.29in}}{{\fontsize{8pt}{9.6pt}\selectfont 0.972}} & 
\multicolumn{1}{|p{0.29in}}{{\fontsize{8pt}{9.6pt}\selectfont 0.9731}} & 
\multicolumn{1}{|p{0.29in}}{{\fontsize{8pt}{9.6pt}\selectfont 0.5556}} & 
\multicolumn{1}{|p{0.29in}}{{\fontsize{8pt}{9.6pt}\selectfont 0.3}} & 
\multicolumn{1}{|p{0.29in}}{{\fontsize{8pt}{9.6pt}\selectfont 1}} & 
\multicolumn{1}{|p{0.29in}|}{{\fontsize{8pt}{9.6pt}\selectfont 1}} \\
\hline
\multicolumn{1}{|p{0.33in}}{{\fontsize{8pt}{9.6pt}\selectfont Average }} & 
\multicolumn{1}{|p{0.3in}}{{\fontsize{8pt}{9.6pt}\selectfont 99.76$\%$ }} & 
\multicolumn{1}{|p{0.3in}}{{\fontsize{8pt}{9.6pt}\selectfont 97.17$\%$ }} & 
\multicolumn{1}{|p{0.29in}}{{\fontsize{8pt}{9.6pt}\selectfont 97.57$\%$ }} & 
\multicolumn{1}{|p{0.29in}}{{\fontsize{8pt}{9.6pt}\selectfont 95.51$\%$ }} & 
\multicolumn{1}{|p{0.29in}}{{\fontsize{8pt}{9.6pt}\selectfont 66.67$\%$ }} & 
\multicolumn{1}{|p{0.29in}}{{\fontsize{8pt}{9.6pt}\selectfont 41.80$\%$ }} & 
\multicolumn{1}{|p{0.29in}}{{\fontsize{8pt}{9.6pt}\selectfont 95.00$\%$ }} & 
\multicolumn{1}{|p{0.29in}|}{{\fontsize{8pt}{9.6pt}\selectfont 92.00$\%$ }} \\
\hline

\end{tabular}
\label{Tbl1}
 \end{table*}

To verify the need for dynamic detection under r-fMRI data for each subject after pre-treatment. We carry out a controlled experiment with data processed by BCCPM and with out it. In both sets,a five-fold cross-validation using LBEM to extract local features were commit, which then classified using KELM. The procedure was repeated 30 times and the average classification result for each fold was put down. The experimental outcomes are shown in Table \ref{Tbl1}. It shows that after dynamic detection, the same classification method bring off better classification accuracy and proves that the brain has temporal dynamics even at rest.

Finally, we assimilate the effects of the different parameters on classification accuracy. We analyse the outcomes with two types of parameters: a) Repeated the whole process with Linearl and RBF kernels. While using these two different kernel functions, for this experiment RBF achieves the higher accuracy. The experimental results are shown in Table \ref{Tbl2}.
\begin{table*}[h!]
 \caption{Average accuracy comparison using different kernels}\label{Tbl2}
 			\centering
\begin{tabular}{p{0.4in}p{0.64in}p{0.8in}p{0.92in}p{0.77in}}
\hline
\multicolumn{1}{|p{0.4in}}{\multirow{1}{*}{\begin{tabular}{p{0.4in}}{\fontsize{8pt}{9.6pt}\selectfont \textbf{Class}}\\\end{tabular}}} & 
\multicolumn{2}{|p{\dimexpr1.65in+2\tabcolsep\relax}}{{\fontsize{8pt}{9.6pt}\selectfont \textbf{KHELM Using RBF kernel}}} & 
\multicolumn{2}{|p{\dimexpr1.9in+2\tabcolsep\relax}|}{{\fontsize{8pt}{9.6pt}\selectfont \textbf{KHELM Using Linear kernel}}} \\
\hline
\multicolumn{1}{|p{0.4in}}{} & 
\multicolumn{1}{|p{0.64in}}{{\fontsize{8pt}{9.6pt}\selectfont \textbf{\textit{NC}}}} & 
\multicolumn{1}{|p{0.8in}}{{\fontsize{8pt}{9.6pt}\selectfont \textbf{\textit{ADHD}}}} & 
\multicolumn{1}{|p{0.92in}}{{\fontsize{8pt}{9.6pt}\selectfont \textbf{\textit{NC}}}} & 
\multicolumn{1}{|p{0.77in}|}{{\fontsize{8pt}{9.6pt}\selectfont \textbf{\textit{ADHD}}}} \\
\hline
\multicolumn{1}{|p{0.4in}}{{\fontsize{8pt}{9.6pt}\selectfont Fold1}} & 
\multicolumn{1}{|p{0.64in}}{{\fontsize{8pt}{9.6pt}\selectfont 1}} & 
\multicolumn{1}{|p{0.8in}}{{\fontsize{8pt}{9.6pt}\selectfont 0.8462}} & 
\multicolumn{1}{|p{0.92in}}{{\fontsize{8pt}{9.6pt}\selectfont 0.9615}} & 
\multicolumn{1}{|p{0.77in}|}{{\fontsize{8pt}{9.6pt}\selectfont 0.6154}} \\
\hline
\multicolumn{1}{|p{0.4in}}{{\fontsize{8pt}{9.6pt}\selectfont Fold2}} & 
\multicolumn{1}{|p{0.64in}}{{\fontsize{8pt}{9.6pt}\selectfont 0.9600}} & 
\multicolumn{1}{|p{0.8in}}{{\fontsize{8pt}{9.6pt}\selectfont 1}} & 
\multicolumn{1}{|p{0.92in}}{{\fontsize{8pt}{9.6pt}\selectfont 0.9600}} & 
\multicolumn{1}{|p{0.77in}|}{{\fontsize{8pt}{9.6pt}\selectfont 0.6154}} \\
\hline
\multicolumn{1}{|p{0.4in}}{{\fontsize{8pt}{9.6pt}\selectfont Fold3}} & 
\multicolumn{1}{|p{0.64in}}{{\fontsize{8pt}{9.6pt}\selectfont 1}} & 
\multicolumn{1}{|p{0.8in}}{{\fontsize{8pt}{9.6pt}\selectfont 1}} & 
\multicolumn{1}{|p{0.92in}}{{\fontsize{8pt}{9.6pt}\selectfont 0.9600}} & 
\multicolumn{1}{|p{0.77in}|}{{\fontsize{8pt}{9.6pt}\selectfont 0.6429}} \\
\hline
\multicolumn{1}{|p{0.4in}}{{\fontsize{8pt}{9.6pt}\selectfont Fold4}} & 
\multicolumn{1}{|p{0.64in}}{{\fontsize{8pt}{9.6pt}\selectfont 1}} & 
\multicolumn{1}{|p{0.8in}}{{\fontsize{8pt}{9.6pt}\selectfont 1}} & 
\multicolumn{1}{|p{0.92in}}{{\fontsize{8pt}{9.6pt}\selectfont 0.9615}} & 
\multicolumn{1}{|p{0.77in}|}{{\fontsize{8pt}{9.6pt}\selectfont 0.6923}} \\
\hline
\multicolumn{1}{|p{0.4in}}{{\fontsize{8pt}{9.6pt}\selectfont Fold5}} & 
\multicolumn{1}{|p{0.64in}}{{\fontsize{8pt}{9.6pt}\selectfont 1}} & 
\multicolumn{1}{|p{0.8in}}{{\fontsize{8pt}{9.6pt}\selectfont 1}} & 
\multicolumn{1}{|p{0.92in}}{{\fontsize{8pt}{9.6pt}\selectfont 0.9600}} & 
\multicolumn{1}{|p{0.77in}|}{{\fontsize{8pt}{9.6pt}\selectfont 0.4615}} \\
\hline
\multicolumn{1}{|p{0.4in}}{{\fontsize{8pt}{9.6pt}\selectfont Average }} & 
\multicolumn{1}{|p{0.64in}}{{\fontsize{8pt}{9.6pt}\selectfont 99.2$\%$ }} & 
\multicolumn{1}{|p{0.8in}}{{\fontsize{8pt}{9.6pt}\selectfont 96.924$\%$ }} & 
\multicolumn{1}{|p{0.92in}}{{\fontsize{8pt}{9.6pt}\selectfont 96.06$\%$ }} & 
\multicolumn{1}{|p{0.77in}|}{{\fontsize{8pt}{9.6pt}\selectfont 60.55$\%$ }} \\
\hline

\end{tabular}
 \end{table*}

As depicted in Table \ref{Tbl2},  that the linear kernel function average ADHD is 60.55\%, average accuracy of NC is 96.06\%), the calculation result of the RBF kernel function (average accuracy of ADHD is 96.92\%, average accuracy of NC is 99.2\%) is more accurate. Here we get better results using RBF kernel, which has been proved by different researchers by comparing different works using different kernels \cite{RF30}\cite{RF31}. b) the effect of increasing number of hidden layers on classification accuracy using KH-ELM. The same method was used for HELM to perform a comparative experiment and the results are shown in Table \ref{Tbl3}. (59.19\% for ADHD and 85.97\% for NC). The experiments show that the classification results of HELM [32] are very unstable compared to the high classification accuracy of KH-ELM. Table \ref{Tbl3} shows the experimental results of KH-ELM with three hidden layers (41.80\% for ADHD and 66.67\% for NC) and with one hidden layer (97.17\% for ADHD and 99.76\% for NC).
 
\begin{table*}[h!]
\caption{The comparison of classification accuracy in different parameters}\label{Tbl3}
 			\centering
\begin{tabular}{p{0.33in}p{0.56in}p{0.56in}p{0.54in}p{0.54in}p{0.29in}p{0.32in}}
\hline
\multicolumn{1}{|p{0.33in}}{\multirow{1}{*}{\begin{tabular}{p{0.33in}}{\fontsize{8pt}{9.6pt}\selectfont \textbf{Class}}\\\end{tabular}}} & 
\multicolumn{2}{|p{\dimexpr1.31in+2\tabcolsep\relax}} {\fontsize{8pt}{9.6pt}\selectfont \textbf{KHELM with Three Hidden Layer}} & 
\multicolumn{2}{|p{\dimexpr1.28in+2\tabcolsep\relax}} {\fontsize{8pt}{9.6pt}\selectfont \textbf{ KHELM with One Hidden Layer}} & 
\multicolumn{2}{|p{\dimexpr0.81in+2\tabcolsep\relax}|} {\fontsize{8pt}{9.6pt}\selectfont \textbf{HELM}} \\
\hline
\multicolumn{1}{|p{0.33in}}{} & 
\multicolumn{1}{|p{0.56in}}{{\fontsize{8pt}{9.6pt}\selectfont \textbf{\textit{NC}}}} & 
\multicolumn{1}{|p{0.56in}}{{\fontsize{8pt}{9.6pt}\selectfont \textbf{\textit{ADHD}}}} & 
\multicolumn{1}{|p{0.54in}}{{\fontsize{8pt}{9.6pt}\selectfont \textbf{\textit{NC}}}} & 
\multicolumn{1}{|p{0.54in}}{{\fontsize{8pt}{9.6pt}\selectfont \textbf{\textit{ADHD}}}} & 
\multicolumn{1}{|p{0.29in}}{{\fontsize{8pt}{9.6pt}\selectfont \textbf{\textit{NC}}}} & 
\multicolumn{1}{|p{0.32in}|}{{\fontsize{8pt}{9.6pt}\selectfont \textbf{\textit{ADHD}}}} \\
\hline
\multicolumn{1}{|p{0.33in}}{{\fontsize{8pt}{9.6pt}\selectfont Fold1}} & 
\multicolumn{1}{|p{0.56in}}{{\fontsize{8pt}{9.6pt}\selectfont 0.7111}} & 
\multicolumn{1}{|p{0.56in}}{{\fontsize{8pt}{9.6pt}\selectfont 0.44}} & 
\multicolumn{1}{|p{0.54in}}{{\fontsize{8pt}{9.6pt}\selectfont 0.9973}} & 
\multicolumn{1}{|p{0.54in}}{{\fontsize{8pt}{9.6pt}\selectfont 1}} & 
\multicolumn{1}{|p{0.29in}}{{\fontsize{8pt}{9.6pt}\selectfont 0.8667}} & 
\multicolumn{1}{|p{0.32in}|}{{\fontsize{8pt}{9.6pt}\selectfont 0.5974}} \\
\hline
\multicolumn{1}{|p{0.33in}}{{\fontsize{8pt}{9.6pt}\selectfont Fold2}} & 
\multicolumn{1}{|p{0.56in}}{{\fontsize{8pt}{9.6pt}\selectfont 0.7778}} & 
\multicolumn{1}{|p{0.56in}}{{\fontsize{8pt}{9.6pt}\selectfont 0.4}} & 
\multicolumn{1}{|p{0.54in}}{{\fontsize{8pt}{9.6pt}\selectfont 0.9907}} & 
\multicolumn{1}{|p{0.54in}}{{\fontsize{8pt}{9.6pt}\selectfont 1}} & 
\multicolumn{1}{|p{0.29in}}{{\fontsize{8pt}{9.6pt}\selectfont 0.892}} & 
\multicolumn{1}{|p{0.32in}|}{{\fontsize{8pt}{9.6pt}\selectfont 0.4692}} \\
\hline
\multicolumn{1}{|p{0.33in}}{{\fontsize{8pt}{9.6pt}\selectfont Fold3}} & 
\multicolumn{1}{|p{0.56in}}{{\fontsize{8pt}{9.6pt}\selectfont 0.6222}} & 
\multicolumn{1}{|p{0.56in}}{{\fontsize{8pt}{9.6pt}\selectfont 0.4}} & 
\multicolumn{1}{|p{0.54in}}{{\fontsize{8pt}{9.6pt}\selectfont 1}} & 
\multicolumn{1}{|p{0.54in}}{{\fontsize{8pt}{9.6pt}\selectfont 0.9974}} & 
\multicolumn{1}{|p{0.29in}}{{\fontsize{8pt}{9.6pt}\selectfont 0.8267}} & 
\multicolumn{1}{|p{0.32in}|}{{\fontsize{8pt}{9.6pt}\selectfont 0.6359}} \\
\hline
\multicolumn{1}{|p{0.33in}}{{\fontsize{8pt}{9.6pt}\selectfont Fold4}} & 
\multicolumn{1}{|p{0.56in}}{{\fontsize{8pt}{9.6pt}\selectfont 0.6667}} & 
\multicolumn{1}{|p{0.56in}}{{\fontsize{8pt}{9.6pt}\selectfont 0.55}} & 
\multicolumn{1}{|p{0.54in}}{{\fontsize{8pt}{9.6pt}\selectfont 1}} & 
\multicolumn{1}{|p{0.54in}}{{\fontsize{8pt}{9.6pt}\selectfont 0.9564}} & 
\multicolumn{1}{|p{0.29in}}{{\fontsize{8pt}{9.6pt}\selectfont 0.8667}} & 
\multicolumn{1}{|p{0.32in}|}{{\fontsize{8pt}{9.6pt}\selectfont 0.6}} \\
\hline
\multicolumn{1}{|p{0.33in}}{{\fontsize{8pt}{9.6pt}\selectfont Fold5}} & 
\multicolumn{1}{|p{0.56in}}{{\fontsize{8pt}{9.6pt}\selectfont 0.5556}} & 
\multicolumn{1}{|p{0.56in}}{{\fontsize{8pt}{9.6pt}\selectfont 0.3}} & 
\multicolumn{1}{|p{0.54in}}{{\fontsize{8pt}{9.6pt}\selectfont 1}} & 
\multicolumn{1}{|p{0.54in}}{{\fontsize{8pt}{9.6pt}\selectfont 0.9048}} & 
\multicolumn{1}{|p{0.29in}}{{\fontsize{8pt}{9.6pt}\selectfont 0.8462}} & 
\multicolumn{1}{|p{0.32in}|}{{\fontsize{8pt}{9.6pt}\selectfont 0.6571}} \\
\hline
\multicolumn{1}{|p{0.33in}}{{\fontsize{8pt}{9.6pt}\selectfont Average }} & 
\multicolumn{1}{|p{0.56in}}{{\fontsize{8pt}{9.6pt}\selectfont 66.67$\%$ }} & 
\multicolumn{1}{|p{0.56in}}{{\fontsize{8pt}{9.6pt}\selectfont 41.80$\%$ }} & 
\multicolumn{1}{|p{0.54in}}{{\fontsize{8pt}{9.6pt}\selectfont 99.76$\%$ }} & 
\multicolumn{1}{|p{0.54in}}{{\fontsize{8pt}{9.6pt}\selectfont 97.17$\%$ }} & 
\multicolumn{1}{|p{0.29in}}{{\fontsize{8pt}{9.6pt}\selectfont 85.97$\%$ }} & 
\multicolumn{1}{|p{0.32in}|}{{\fontsize{8pt}{9.6pt}\selectfont 59.19$\%$ }} \\
\hline

\end{tabular}
 \end{table*}

We have found that the accuracy of the classification results decreases as the number of hidden levels increases. To illustrate more, identical operation was carry out with up to six hidden layers. Same protocol has being followed as above. The experiment was repeated 30 times and record the average result of each convolution, gradually increasing the number of layers. The details are shown in Table \ref{Tbl4} and \ref{Tbl5}.

\begin{table*}[h!]
\caption{Accuracy comparison of KH-ELM with different no of hidden layers (1-3)}\label{Tbl4}
 			\centering
\begin{tabular}{p{0.4in}p{0.48in}p{0.36in}p{0.48in}p{0.55in}p{0.46in}p{0.37in}}
\hline
\multicolumn{1}{|p{0.4in}}{\multirow{1}{*}{\begin{tabular}{p{0.4in}}{\fontsize{8pt}{9.6pt}\selectfont \textbf{Class}}\\\end{tabular}}} & 
\multicolumn{2}{|p{\dimexpr1.04in+2\tabcolsep\relax}}{{\fontsize{8pt}{9.6pt}\selectfont \textbf{ KH-ELM with One Hidden Layer}}} & 
\multicolumn{2}{|p{\dimexpr1.23in+2\tabcolsep\relax}}{{\fontsize{8pt}{9.6pt}\selectfont \textbf{KH-ELM with second Hidden Layer}}} & 
\multicolumn{2}{|p{\dimexpr1.03in+2\tabcolsep\relax}|}{{\fontsize{8pt}{9.6pt}\selectfont \textbf{KH-ELM with Third Hidden Layer}}} \\
\hline
\multicolumn{1}{|p{0.4in}}{} & 
\multicolumn{1}{|p{0.48in}}{{\fontsize{8pt}{9.6pt}\selectfont \textbf{\textit{NC}}}} & 
\multicolumn{1}{|p{0.36in}}{{\fontsize{8pt}{9.6pt}\selectfont \textbf{\textit{ADHD}}}} & 
\multicolumn{1}{|p{0.48in}}{{\fontsize{8pt}{9.6pt}\selectfont \textbf{\textit{NC}}}} & 
\multicolumn{1}{|p{0.55in}}{{\fontsize{8pt}{9.6pt}\selectfont \textbf{\textit{ADHD}}}} & 
\multicolumn{1}{|p{0.46in}}{{\fontsize{8pt}{9.6pt}\selectfont \textbf{\textit{NC}}}} & 
\multicolumn{1}{|p{0.37in}|}{{\fontsize{8pt}{9.6pt}\selectfont \textbf{\textit{ADHD}}}} \\
\hline
\multicolumn{1}{|p{0.4in}}{{\fontsize{8pt}{9.6pt}\selectfont Fold1}} & 
\multicolumn{1}{|p{0.48in}}{{\fontsize{8pt}{9.6pt}\selectfont 0.9973}} & 
\multicolumn{1}{|p{0.36in}}{{\fontsize{8pt}{9.6pt}\selectfont 1}} & 
\multicolumn{1}{|p{0.48in}}{{\fontsize{8pt}{9.6pt}\selectfont 0.9423}} & 
\multicolumn{1}{|p{0.55in}}{{\fontsize{8pt}{9.6pt}\selectfont 0.9025}} & 
\multicolumn{1}{|p{0.46in}}{{\fontsize{8pt}{9.6pt}\selectfont 0.7111}} & 
\multicolumn{1}{|p{0.37in}|}{{\fontsize{8pt}{9.6pt}\selectfont 0.44}} \\
\hline
\multicolumn{1}{|p{0.4in}}{{\fontsize{8pt}{9.6pt}\selectfont Fold2}} & 
\multicolumn{1}{|p{0.48in}}{{\fontsize{8pt}{9.6pt}\selectfont 0.9907}} & 
\multicolumn{1}{|p{0.36in}}{{\fontsize{8pt}{9.6pt}\selectfont 1}} & 
\multicolumn{1}{|p{0.48in}}{{\fontsize{8pt}{9.6pt}\selectfont 0.972}} & 
\multicolumn{1}{|p{0.55in}}{{\fontsize{8pt}{9.6pt}\selectfont 0.7974}} & 
\multicolumn{1}{|p{0.46in}}{{\fontsize{8pt}{9.6pt}\selectfont 0.7778}} & 
\multicolumn{1}{|p{0.37in}|}{{\fontsize{8pt}{9.6pt}\selectfont 0.4}} \\
\hline
\multicolumn{1}{|p{0.4in}}{{\fontsize{8pt}{9.6pt}\selectfont Fold3}} & 
\multicolumn{1}{|p{0.48in}}{{\fontsize{8pt}{9.6pt}\selectfont 1}} & 
\multicolumn{1}{|p{0.36in}}{{\fontsize{8pt}{9.6pt}\selectfont 0.9974}} & 
\multicolumn{1}{|p{0.48in}}{{\fontsize{8pt}{9.6pt}\selectfont 0.976}} & 
\multicolumn{1}{|p{0.55in}}{{\fontsize{8pt}{9.6pt}\selectfont 0.9642}} & 
\multicolumn{1}{|p{0.46in}}{{\fontsize{8pt}{9.6pt}\selectfont 0.6222}} & 
\multicolumn{1}{|p{0.37in}|}{{\fontsize{8pt}{9.6pt}\selectfont 0.4}} \\
\hline
\multicolumn{1}{|p{0.4in}}{{\fontsize{8pt}{9.6pt}\selectfont Fold4}} & 
\multicolumn{1}{|p{0.48in}}{{\fontsize{8pt}{9.6pt}\selectfont 1}} & 
\multicolumn{1}{|p{0.36in}}{{\fontsize{8pt}{9.6pt}\selectfont 0.9564}} & 
\multicolumn{1}{|p{0.48in}}{{\fontsize{8pt}{9.6pt}\selectfont 0.9667}} & 
\multicolumn{1}{|p{0.55in}}{{\fontsize{8pt}{9.6pt}\selectfont 0.9359}} & 
\multicolumn{1}{|p{0.46in}}{{\fontsize{8pt}{9.6pt}\selectfont 0.6667}} & 
\multicolumn{1}{|p{0.37in}|}{{\fontsize{8pt}{9.6pt}\selectfont 0.55}} \\
\hline
\multicolumn{1}{|p{0.4in}}{{\fontsize{8pt}{9.6pt}\selectfont Fold5}} & 
\multicolumn{1}{|p{0.48in}}{{\fontsize{8pt}{9.6pt}\selectfont 1}} & 
\multicolumn{1}{|p{0.36in}}{{\fontsize{8pt}{9.6pt}\selectfont 0.9048}} & 
\multicolumn{1}{|p{0.48in}}{{\fontsize{8pt}{9.6pt}\selectfont 0.9586}} & 
\multicolumn{1}{|p{0.55in}}{{\fontsize{8pt}{9.6pt}\selectfont 0.8846}} & 
\multicolumn{1}{|p{0.46in}}{{\fontsize{8pt}{9.6pt}\selectfont 0.5556}} & 
\multicolumn{1}{|p{0.37in}|}{{\fontsize{8pt}{9.6pt}\selectfont 0.3}} \\
\hline
\multicolumn{1}{|p{0.4in}}{{\fontsize{8pt}{9.6pt}\selectfont Average }} & 
\multicolumn{1}{|p{0.48in}}{{\fontsize{8pt}{9.6pt}\selectfont 99.76$\%$ }} & 
\multicolumn{1}{|p{0.36in}}{{\fontsize{8pt}{9.6pt}\selectfont 97.17$\%$ }} & 
\multicolumn{1}{|p{0.48in}}{{\fontsize{8pt}{9.6pt}\selectfont 96.31$\%$ }} & 
\multicolumn{1}{|p{0.55in}}{{\fontsize{8pt}{9.6pt}\selectfont 89.69$\%$ }} & 
\multicolumn{1}{|p{0.46in}}{{\fontsize{8pt}{9.6pt}\selectfont 66.67$\%$ }} & 
\multicolumn{1}{|p{0.37in}|}{{\fontsize{8pt}{9.6pt}\selectfont 41.80$\%$ }} \\
\hline

\end{tabular}
 \end{table*}

\begin{table*}[h!]
\caption{Accuracy comparison of KH-ELM with different no of hidden layers (4-6}\label{Tbl5}
 			\centering
\begin{tabular}{p{0.38in}p{0.46in}p{0.41in}p{0.49in}p{0.53in}p{0.48in}p{0.36in}}
\hline
\multicolumn{1}{|p{0.38in}}{\multirow{1}{*}{\begin{tabular}{p{0.38in}}{\fontsize{8pt}{9.6pt}\selectfont \textbf{Class}}\\\end{tabular}}} & 
\multicolumn{2}{|p{\dimexpr1.07in+2\tabcolsep\relax}}{{\fontsize{8pt}{9.6pt}\selectfont \textbf{KH-ELM with Forth Hidden Layer}}} & 
\multicolumn{2}{|p{\dimexpr1.21in+2\tabcolsep\relax}}{{\fontsize{8pt}{9.6pt}\selectfont \textbf{KH-ELM with Fifth Hidden Layer}}} & 
\multicolumn{2}{|p{\dimexpr1.04in+2\tabcolsep\relax}|}{{\fontsize{8pt}{9.6pt}\selectfont \textbf{KH-ELM with Sixth Hidden Layer}}} \\
\hline
\multicolumn{1}{|p{0.38in}}{} & 
\multicolumn{1}{|p{0.46in}}{{\fontsize{8pt}{9.6pt}\selectfont \textbf{\textit{NC}}}} & 
\multicolumn{1}{|p{0.41in}}{{\fontsize{8pt}{9.6pt}\selectfont \textbf{\textit{ADHD}}}} & 
\multicolumn{1}{|p{0.49in}}{{\fontsize{8pt}{9.6pt}\selectfont \textbf{\textit{NC}}}} & 
\multicolumn{1}{|p{0.53in}}{{\fontsize{8pt}{9.6pt}\selectfont \textbf{\textit{ADHD}}}} & 
\multicolumn{1}{|p{0.48in}}{{\fontsize{8pt}{9.6pt}\selectfont \textbf{\textit{NC}}}} & 
\multicolumn{1}{|p{0.36in}|}{{\fontsize{8pt}{9.6pt}\selectfont \textbf{\textit{ADHD}}}} \\
\hline
\multicolumn{1}{|p{0.38in}}{{\fontsize{8pt}{9.6pt}\selectfont Fold1}} & 
\multicolumn{1}{|p{0.46in}}{{\fontsize{8pt}{9.6pt}\selectfont 0.8307}} & 
\multicolumn{1}{|p{0.41in}}{{\fontsize{8pt}{9.6pt}\selectfont 0.3538}} & 
\multicolumn{1}{|p{0.49in}}{{\fontsize{8pt}{9.6pt}\selectfont 0.7474}} & 
\multicolumn{1}{|p{0.53in}}{{\fontsize{8pt}{9.6pt}\selectfont 0.3743}} & 
\multicolumn{1}{|p{0.48in}}{{\fontsize{8pt}{9.6pt}\selectfont 0.7397}} & 
\multicolumn{1}{|p{0.36in}|}{{\fontsize{8pt}{9.6pt}\selectfont 0.3512}} \\
\hline
\multicolumn{1}{|p{0.38in}}{{\fontsize{8pt}{9.6pt}\selectfont Fold2}} & 
\multicolumn{1}{|p{0.46in}}{{\fontsize{8pt}{9.6pt}\selectfont 0.7693}} & 
\multicolumn{1}{|p{0.41in}}{{\fontsize{8pt}{9.6pt}\selectfont 0.5230}} & 
\multicolumn{1}{|p{0.49in}}{{\fontsize{8pt}{9.6pt}\selectfont 0.688}} & 
\multicolumn{1}{|p{0.53in}}{{\fontsize{8pt}{9.6pt}\selectfont 0.5025}} & 
\multicolumn{1}{|p{0.48in}}{{\fontsize{8pt}{9.6pt}\selectfont 0.6973}} & 
\multicolumn{1}{|p{0.36in}|}{{\fontsize{8pt}{9.6pt}\selectfont 0.5051}} \\
\hline
\multicolumn{1}{|p{0.38in}}{{\fontsize{8pt}{9.6pt}\selectfont Fold3}} & 
\multicolumn{1}{|p{0.46in}}{{\fontsize{8pt}{9.6pt}\selectfont 0.7213}} & 
\multicolumn{1}{|p{0.41in}}{{\fontsize{8pt}{9.6pt}\selectfont 0.6278}} & 
\multicolumn{1}{|p{0.49in}}{{\fontsize{8pt}{9.6pt}\selectfont 0.7026}} & 
\multicolumn{1}{|p{0.53in}}{{\fontsize{8pt}{9.6pt}\selectfont 0.5380}} & 
\multicolumn{1}{|p{0.48in}}{{\fontsize{8pt}{9.6pt}\selectfont 0.7346}} & 
\multicolumn{1}{|p{0.36in}|}{{\fontsize{8pt}{9.6pt}\selectfont 0.5166}} \\
\hline
\multicolumn{1}{|p{0.38in}}{{\fontsize{8pt}{9.6pt}\selectfont Fold4}} & 
\multicolumn{1}{|p{0.46in}}{{\fontsize{8pt}{9.6pt}\selectfont 0.8359}} & 
\multicolumn{1}{|p{0.41in}}{{\fontsize{8pt}{9.6pt}\selectfont 0.5256}} & 
\multicolumn{1}{|p{0.49in}}{{\fontsize{8pt}{9.6pt}\selectfont 0.7538}} & 
\multicolumn{1}{|p{0.53in}}{{\fontsize{8pt}{9.6pt}\selectfont 0.4461}} & 
\multicolumn{1}{|p{0.48in}}{{\fontsize{8pt}{9.6pt}\selectfont 0.7667}} & 
\multicolumn{1}{|p{0.36in}|}{{\fontsize{8pt}{9.6pt}\selectfont 0.4410}} \\
\hline
\multicolumn{1}{|p{0.38in}}{{\fontsize{8pt}{9.6pt}\selectfont Fold5}} & 
\multicolumn{1}{|p{0.46in}}{{\fontsize{8pt}{9.6pt}\selectfont 0.724}} & 
\multicolumn{1}{|p{0.41in}}{{\fontsize{8pt}{9.6pt}\selectfont 0.5820}} & 
\multicolumn{1}{|p{0.49in}}{{\fontsize{8pt}{9.6pt}\selectfont 0.672}} & 
\multicolumn{1}{|p{0.53in}}{{\fontsize{8pt}{9.6pt}\selectfont 0.5615}} & 
\multicolumn{1}{|p{0.48in}}{{\fontsize{8pt}{9.6pt}\selectfont 0.644}} & 
\multicolumn{1}{|p{0.36in}|}{{\fontsize{8pt}{9.6pt}\selectfont 0.4897}} \\
\hline
\multicolumn{1}{|p{0.38in}}{{\fontsize{8pt}{9.6pt}\selectfont Average }} & 
\multicolumn{1}{|p{0.46in}}{{\fontsize{8pt}{9.6pt}\selectfont 77.62$\%$ }} & 
\multicolumn{1}{|p{0.41in}}{{\fontsize{8pt}{9.6pt}\selectfont 52.24$\%$ }} & 
\multicolumn{1}{|p{0.49in}}{{\fontsize{8pt}{9.6pt}\selectfont 71.27$\%$ }} & 
\multicolumn{1}{|p{0.53in}}{{\fontsize{8pt}{9.6pt}\selectfont 48.45$\%$ }} & 
\multicolumn{1}{|p{0.48in}}{{\fontsize{8pt}{9.6pt}\selectfont 71.64$\%$ }} & 
\multicolumn{1}{|p{0.36in}|}{{\fontsize{8pt}{9.6pt}\selectfont 46.07$\%$ }} \\
\hline

\end{tabular}
 \end{table*}
As the deep learning methods generally produce better results when the number of hidden layers increases. But in our experiments, we obtained worse results when we increased the number of layers. The reason for this conclusion is that in experiments with multiple layers, some features are extracted after each layer, which leads to the repetition of attributes, and more features are generated simultaneously. These features are incompatible with the proposed algorithm and have the most significant impact on the outputs when the number of layers increases because the extension layer is limited. After that, we no longer extract features but tend to "overfit" the data, and overfitting reduces the accuracy of the experiment. Therefore, the KH-ELM classification with only one hidden layer for feature extraction offers more efficiency than the one with multiple hidden layers.
\section{Conclusion}
In this paper, we propose a novel classification scheme to improve the diagnosis accuracy from brain imaging data to distinguish the ADHD in comparison to the NC from the brain imaging data of patients. The innovation of our work is to use a hierarchical ELM sparse autoencoder to extract features prior to KELM, which outperformed other classifiers concerning the accuracy rate while requiring fewer hidden layers. Our experiments proved that the proposed framework offers superior efficiency regarding the classification metrics; however, the dynamic detection and selection of the number of layers for classifier feature extraction is not as good as possible.
We try to extent our study by conducting further experiments in the future to improve the average accuracy of fMRI classification for discriminating between the ADHD and NC diseases.

\section{Acknowledgment}

The authors would like to thank Prof.Tianming Liu and his group CAID for providing the ADHD data.

\printcredits

\end{document}